# Structure and Parameter Learning for Causal Independence and Causal Interaction Models

Christopher Meek and David Heckerman
Microsoft Research
Redmond WA, 98052-6399
{meek,heckerma}@microsoft.com

## Abstract

We begin by discussing causal independence models and generalize these models to causal interaction models. Causal interaction models are models that have independent mechanisms where mechanisms can have several causes. In addition to introducing several particular types of causal interaction models, we show how we can apply the Bayesian approach to learning causal interaction models obtaining approximate posterior distributions for the models and obtain MAP and ML estimates for the parameters. We illustrate the approach with a simulation study of learning model posteriors.

## 1 Introduction

Models of causal independence[1] such as the Noisy-or (Good, 1961; Kim and Pearl, 1983) and Noisy-Max (Henrion, 1987) have proved to be useful for probabilistic assessment (Pearl, 1988; Henrion, 1987; Heckerman and Breese, 1996). In addition to easier assessment, there are techniques for performing inference efficiently in models with causal independence (e.g., Heckerman and Breese, 1996; Zhang and Poole, 1996) and techniques to efficiently calculate upper and lower bounds for likelihoods where exact inference is intractable (Jaakkola and Jordan, 1996). The essential idea of causal independence models is that the causes lead to the effect through independent mechanisms. If this type of model is assumed then one only needs to separately assess the probability distributions that describes a mechanism and give a rule for combining the results of the mechanisms. On the other hand, when using full probability tables to represent the conditional distribution of the effect given its causes, we are essentially allowing for *complete causal interactions* between the causes.

The first part of this paper introduces *causal interaction models*. Like the causal independence model, a causal interaction model is a set of mechanisms, a set of causes, and an effect. Unlike the causal independence model, a cause need not be associated with a single mechanism and multiple causes can be associated with a single mechanism. Allowing several causes to be associated with a single mechanism allows for *partial causal interaction* between a set of causes, thus, causal interaction models generalize both the causal independence model and the complete causal interaction model. In Section 2, we show how to represent causal interaction models as directed acyclic graphical (DAG) models (a.k.a., Bayesian networks, belief networks, causal networks) with hidden variables. In addition we introduce a special type of causal interaction model, the *exponential causal interaction model*. Examples of exponential causal interaction models are given in Section 3.

In the second part of the paper we turn our attention from representation to learning the structure and parameters of exponential causal interaction models. In much of the initial work on learning (discrete) DAG models, the focus was on learning the structure of the network assuming there were *full conditional probability tables* for each variable in the network. The conditional probability table for a variable represented the conditional probability of the variable given every possible combination of the values of its parents in the DAG model structure. In this representation, the number of parameters associated with a variable is exponential in the number of parents of the variable. This exponential explosion can restrict the set of network structures that can be learned by some methods (e.g., MDL methods, Boukaert 1995). In part, because of these limitations, there has been interest in learning DAG models with more parsimonious representations for the conditional probability of variables given their parents. For instance in Friedman and Goldszmidt (1996) and Chickering et al. (1997), the authors consider using decision trees and a generalization of decision trees to represent the conditional probability of the variable given its parents. These representations of *local structure* allows for dramatic reductions in the dimension of the parameter space. Causal interaction models provide an alternative representation for the local structure in a DAG model. We illustrate the fact that there are Noisy-Max-Interaction models that can not be parsimoniously represented by decision trees and that decision trees and other types of local structures can be embedded in causal interaction models. Thus, causal interaction models are rich set of models for parsimoniously representing local structure.

---

[1] Causal independence is sometimes referred to as intercausal independence.

Since causal interaction models are DAG models with hidden variables and hidden variables are just the extreme case of missing data we discuss learning DAG models with missing data in Section 4. We also discuss how one can use the EM algorithm to obtain ML and MAP estimates for hidden variable models. Finally, in Section 5, we illustrate the fact that one can learn the structure of causal interaction models in a small simulation study. In addition, Section 5 illustrates the importance of correctly calculating the dimension of hidden variable models when learning structure.

## 2  Causal Independence and Causal Interaction models

When constructing a parameterized DAG models, one must specify the conditional probability of each variable given each possible configuration of the parents. Figure 1a shows a variable $E$ with several parents (causes). It is often not feasible to specify a complete probability table to represent the required probabilities, because the number of probabilities grows exponentially in the number of parents. In addition, several authors have argued that this model is inaccurate because it fails to represent the independence of causal interactions.

To overcome both of these inadequacies, researchers have used DAG models such as the one shown in Figure 1b to represent causal independence (e.g., Good, 1961; Kim and Pearl, 1983; Henrion, 1987; Srinivas, 1992). We shall call the $C_i$'s the causes, $E$ the effect, and the $X_i$'s the *noisy mechanism variables*. The (noisy) mechanism variable $X_i$ represents the contribution of the $i^{th}$ mechanism to the effect $E$ where the value of $E$ is a deterministic function (indicated by the double circle in the graph) of the values of the mechanism variables. The independence of the causal mechanisms is captured by (1) the conditional independence of the mechanism variables given the causes, and (2) the independence between the set of mechanism variables for $E$ and other variables in the network (not depicted in Figure 1b) given the causes and the effect.

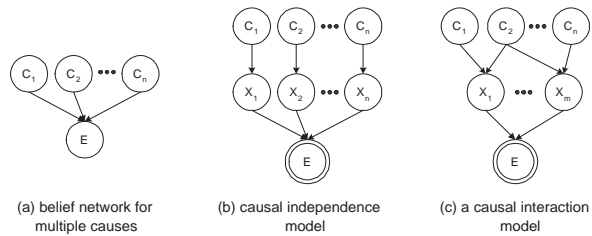

Figure 1: Different types of local structure.

A causal interaction model relaxes the restrictions that each cause has a unique mechanism variable and that each mechanism variable has a unique cause. Figure 1c shows an example of a causal interaction model. With a causal interaction model, it is possible to model relationships in which some of the causes interact to cause the effect and some of the causes act independently. Example of interactions are often found in medicine. For instance, in some studies *smoking* and *estrogen level* have been found to have a synergistic effect on the rate of stroke in females. There is no reason to stop the modeling of the causal process at this level. The $i^{th}$ mechanism described by the conditional distribution of $X_i$ given the parent of $X_i$ could be modeled as a decision tree, or a model with additional hidden variables.

Roughly, a mechanism describes one "path" through which a set of causes lead to an effect. A *mechanism* for causes $C_1, \ldots, C_n$ and effect $E$ are a set of nodes $M$ which are not observed (hidden) such that (1) there is a distinguished variable called the *noisy mechanism variable* (or, simply, the mechanism variable), (2) only members of the mechanism $M$ and causes can point to members of $M$, (3) the nodes in $M$ form a directed acyclic graph, (4) the only variable in $M$ that points to a non-member of $M$ is the mechanism variable which only points to $E$. The Figure 2b illustrates and example of a mechanism. Note that a cause can point to multiple nodes in a mechanism.

A causal interaction model is roughly a DAG model of mechanisms which describes the conditional distribution of the effect given its causes. More precisely, a *causal interaction model* is a (1) a set of causes $C_1, \ldots, C_n$, (2) an effect variable $E$, (3) a set of mechanisms for $C_1, \ldots, C_n$ and effect $E$, which we denote by $M_1, \ldots, M_m$, (4) where the value of the effect variable is a deterministic function of the mechanism variables $X_1, \ldots, X_m$, which we call the *combination function*.

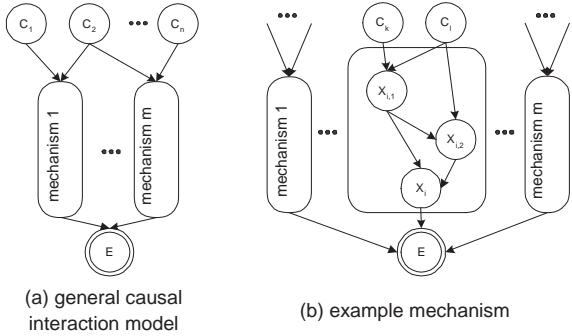

Figure 2: Causal interaction models.

Let $\mathcal{M}$ be the set of all of the variables in mechanisms for causes $C_1, \ldots, C_n$ and effect $E$. As in the case of causal independence models, the independence of the causal mechanisms is captured by (1) the conditional independence of the set of variables in each mechanism given the causes (i.e., for $i \neq j$, $M_i$ is independent of $M_j$ given $C_1, \ldots, C_n$), and (2) the independence between the set of all mechanism variables ($\mathcal{M}$) and other variables in the DAG model given the causes $C_i$ and effect $E$.

It is common to add a leak term to the noisy-or and noisy-max models. A leak term is added to model mechanisms not associated with other variables in the model. A leak term corresponds to a mechanism variable (and thus a mechanism) which does not have any causes that are in the DAG model.

Finally, an *exponential causal interaction model* is a causal interaction model in which the conditional likelihood for each variable in each mechanism is in the exponential family. In Section 3, we discuss a variety of specific exponential causal interaction models. We focus on exponential causal interaction models because with these models we can often find tractable algorithms for inference and with tractable models for inference we can apply the EM algorithm.

## 3 Examples of Exponential Causal Interaction models

In this section we give a few examples of exponential causal interaction models.

### 3.1 Noisy-Max-Interaction models

A *noisy-max-interaction (NMI) model* is a causal interaction model in which, (1) each mechanism consists of a single mechanism variable which has a domain that is a subset of the domain of the effect variable, (2) the domain of the effect variable can be ordered by a binary relation $\leq$, (3) the likelihood of each mechanism variable given the values of its parents is in the exponential family, and (4) the combination function is $max_{\leq}(x_1, \ldots, x_m)$. Note that the effect and the mechanism variables need not be discrete. It follows from the combination function that

$$p(E \leq e | \vec{C} = \vec{c}, \boldsymbol{\theta}) = \prod_{i=1}^{m} p(X_i \leq e | \vec{C} = \vec{c}, \boldsymbol{\theta}).$$

An NMI model in which there is only one cause per mechanism variable is a generalization of the Noisy-or and Noisy-max models. These NMI models are noisy-max models without a distinguished state (e.g., "absent" or "normal"). Of course, one can create a *Noisy-Max-Interaction model with distinguished states* by simply distinguishing one parent configuration for each mechanism variable and forcing the associated parameters to 0 and 1. Clearly, when one fixes parameters one is reducing the number of free parameters in the model. One benefit of models without distinguished states is that they can be easier to learn. In the case where one does not know the distinguished states for each of the mechanism variables, we have an additional learning problem; namely we need to identify which parent configurations are the distinguished states. Of course, if we do know which parent configuration is the distinguished state then we can force the parameter restrictions and use the EM algorithm to calculate the ML or MAP estimate of the parameters and approximate the posteriors on the models.

As a special case we consider a *discrete NMI model*, an NMI in which (1) $E$ is a discrete random variable (not necessarily finite), and (2) each mechanism contains only a mechanism variable. Let $\theta_{ijk} = p(X_i = k | \vec{C} = \vec{c}, \boldsymbol{\theta}) = p(X_i = k | \mathbf{Pa}_{X_i} = j, \boldsymbol{\theta})$. Where $\mathbf{Pa}_{X_i}$ is the set of parents of $X_i$. Thus $p(X_i \leq x_i | \vec{C} = \vec{c}, \boldsymbol{\theta}) = \sum_{k \leq x_i} \theta_{ijk}$. Let $j_i$ be the instantiation of causes for the $i^{th}$ mechanism variable. As discussed in Section 4.1, to use the EM algorithm we will need to calculate $p(X_i = k, \mathbf{Pa}_{X_i} = j | \vec{C} = \vec{c}, E = e, \boldsymbol{\theta}) = I(j = j_i) p(X_i = k | \vec{C} = \vec{c}, E = e, \boldsymbol{\theta})$, where $I(j = j_i)$ is an indicator function that is one if and only if $j = j_i$.

$$p(Xi = k | \vec{C} = \vec{c}, E = e, \boldsymbol{\theta}) = \begin{cases} 0 & k > e \\ \frac{\theta_{ij_ik}(\prod_{i \neq j} p(X_i \leq e | \vec{C} = \vec{c}, \boldsymbol{\theta}) - \prod_{i \neq j} p(X_i < e | \vec{C} = \vec{c}, \boldsymbol{\theta}))}{\prod_{i=1}^{n} p(X_i \leq e | \vec{C} = \vec{c}, \boldsymbol{\theta}) - \prod_{i=1}^{n} p(X_i < e | \vec{C} = \vec{c}, \boldsymbol{\theta})} & k < e \\ 1 - \sum_{l < k} p(X_i = l | \vec{C} = \vec{c}, E = e, \boldsymbol{\theta}) & k = e \end{cases}$$

Note that for each mechanism variable we only need to calculate $p(X_i | \vec{C} = \vec{c}, \boldsymbol{\theta})$. If the conditional distribution is in the exponential family then it is easy to apply the EM algorithm, e.g., if the conditional distribution $p(X_i|\vec{C} = \vec{c}, \boldsymbol{\theta})$ is distributed according to a Poisson or multinomial distribution. In addition, we do not need to have a unique conditional distribution for each instantiation of the parents of the mechanism variable. Rather, one can use a decision tree or a decision graph to reduce the number of conditional distributions and thus reduce the number of parameters needed for specifying the conditional distribution of the mechanism variable. This can even be done when the conditional distribution function is the Poisson distribution. Since the conditional distribution of a mechanism variable can be represented with a decision tree, the NMI model is at least as representationally rich as decision trees.

A noisy-or model (a special case of an NMI model) with $n$ binary causes and a binary effect has $n$ parameters. However, for almost all values of the parameters (all but a set of Lebesgue measure zero) a full probability table, i.e., a complete decision tree, must be used to represent the distribution exactly. Thus, causal interaction models provide a rich representation for modeling conditional distributions. Causal interaction models can be viewed as an alternative to decision trees or decision graphs for parsimonious local representations, however, since decision trees and graphs can be embedded in causal interaction models, they are strictly richer representation. The caveat, as we shall see in Section 4, is that one must use iterative methods in approximating several quantities of interest when using causal interaction models. Under suitable assumptions, this is not the case for decision trees and decision graphs.

Finally, in Noisy-or and Noisy-Max models it is common to add a leak term to model mechanisms not associated with the other variables in the model. As discussed in Section 2, we can add leaks to NMI models, however, the extra degrees of freedom in a NMI model as compared to a Noisy-Max can act somewhat like a leak term in a Noisy-max model.

### 3.2 Noisy-Additive-Interaction models

A *Noisy-Additive-interaction (NAI) model* is a causal interaction model in which, (1) each mechanism consists of a single mechanism variable which has a domain that is a subset of the domain of the effect variable, (2) the domain of the effect variable is closed under addition, (3) the likelihood of each mechanism variable given the values of its parents is in the exponential family, and (4) the combination function is addition, $\sum_{i=1}^{m} X_i$.

As a special case of an NAI model in which the effect is not continuous, we consider a Poisson NAI model. A Poisson NAI model is an NAI model in which (1) $p(X_i|\mathbf{Pa}_{X_i} = j, \boldsymbol{\theta}) = Poisson(\lambda_{(i,j)})$ that is $p(X_i = x|\mathbf{Pa}_{X_i} = j, \boldsymbol{\theta}) = exp(-\lambda_{(i,j)})\frac{\lambda_{(i,j)}^x}{x!}$, and (2) each mechanism contains only a mechanism variable. $\lambda$ is called the rate parameter for the Poisson. In this case $\lambda_{(i,j)}$ is a conditional rate parameter. Let $\lambda_{(i,j)}$ be the parameter for the $i^{th}$ mechanism variable when the parents of the $i^{th}$ mechanism variable are in the $j^{th}$ state. Let $j_i$ be the instantiation of parents of the $i^{th}$ mechanism variable. Using the theorem that the sum of $n$ independent random variables having Poisson distributions with parameters $\lambda_1, \ldots, \lambda_n$ is distributed as a Poisson random variable with parameter $\lambda_1 + \ldots + \lambda_n$ we can characterize the Poisson NAI model with the equation $p(E|\vec{C} = \vec{c}, \boldsymbol{\theta}) = Poisson(\sum_{i=1}^{m} \lambda_{(i,j_i)})$.

Poisson random variables are useful in analyzing rates, e.g., number of web page hits per week or number of headaches per week. Thus, the Poisson NAI model has potential for modeling conditional rates even in cases where the causes of the rate can interact. As with Noisy-Max-Interaction models, for a given mechanism variable, one need not have a unique parameter for each instantiation of the parents of the mechanism variable. Rather, one can use decision trees and decision graphs to reduce the number of parameters needed for specifying the conditional distribution of the mechanism variable.

One interesting feature of the Poisson NAI model is that it is possible to run inference using a clique-tree type inference algorithm despite the fact that the clique potentials are infinite. The trick is to form the clique potentials only after the value of $E$ is known. With the value of $E$ known we can bound the values of the $X_i$'s and thus bound the size of the clique potential.

Let $j_i$ be the instantiation of causes for the $i^{th}$ mechanism variable. As discussed in Section 4.1, to use the EM algorithm we will need to calculate $p(X_i = k, \mathbf{Pa}_{X_i} = j|\vec{C} = \vec{c}, E = e, \boldsymbol{\theta}) = I(j = j_i)p(X_i = k|\vec{C} = \vec{c}, E = e, \boldsymbol{\theta})$, where $I(j = j_i)$ is an indicator function that is one if and only if $j = j_i$. Below is the equation for $p(X1 = k|\vec{C} = \vec{c}, E = e, \boldsymbol{\theta})$ where there are $m$ mechanism variables. Let $\theta_{ijk} = p(X_i = k|\mathbf{Pa}_{X_i} = j, \boldsymbol{\theta})$. The inferences for other mechanism variables are analogous.

$$p(X1 = k|\vec{C} = \vec{c}, E = e, \boldsymbol{\theta}) = \begin{cases} 0 & k > e \\ \frac{\theta_{ij_ik}(\sum_{l_2=0}^{e-k} \cdots \sum_{l_m=0}^{e-k-\sum_i l_i} \prod_{o=2}^{n} \theta_{oj_il_i})}{\sum_{l_1=0}^{e} \cdots \sum_{l_m=0}^{e-k-\sum_i l_i} \prod_{o=1}^{n} \theta_{oj_il_i}} & k \leq e \end{cases}$$

A case where inference is even easier are Gaussian NAI

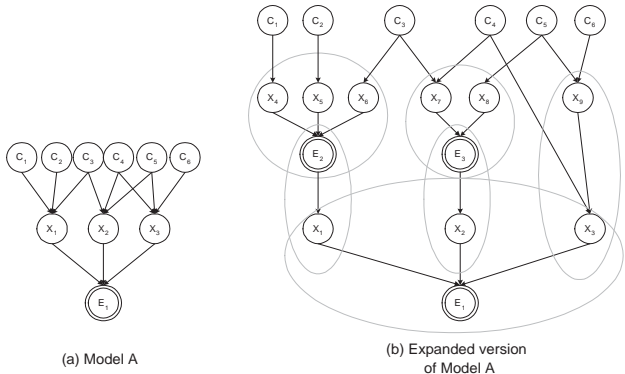

(a) Model A  (b) Expanded version of Model A

Figure 3: Interaction model with nested structure and conditional clique tree.

models. A Gaussian NAI model is a causal interaction model in which the conditional distribution of each of the mechanism variables is Gaussian. By including a discrete finite state hidden variable inside a mechanism it is possible to have conditional distribution for the mechanism variables which are mixtures of Gaussians. In other types of NAI models, e.g., where some of the conditional distribution are Gaussian and some Poisson, inference is more difficult.

### 3.3 Other models

Both the NMI and NAI models have fairly simple structure. Figure 3b illustrates a causal interaction model with a more complicated nested structure. As with any causal interaction model, there is a layer of mechanism nodes followed by a deterministic combination function. The expanded version of Model A in Figure 3b illustrates that the conditional distribution of the mechanism nodes given its parent causes can have nested structure. In this case, the mechanisms associated mechanism variables $X_1$ and $X_2$ have nested causal interaction models and the mechanism associated with mechanism variable $X_3$ has a nested hidden variable $X_9$. It is important to note that when the values of $E$ and the $C_i$'s are observed all of hidden variables in the interaction model are d-separated from other variables in the model, i.e., variables not in the interaction model, and thus inference for EM can be localized to the interaction model.

One might think that inference and thus using EM would be computationally hopeless in the expanded version of Model A in Figure 3b or more complicated causal interaction models. This is not always the case. For the expanded version of Model A the interaction structure conditional on the $C_i$'s forms a polytree. Thus the polynomial-time algorithm of Kim and Pearl (1983) can be used for inference. More generally, the independence of the mechanisms in a causal interaction model lead to computational efficiencies in inference because, in the clique tree conditional on the $C_i$'s, the nodes from different mechanism are only connected by paths through mechanism variables. This point is illustrated by the conditional clique tree in Figure 3b.

In addition to allowing for nested structure, causal interaction models also allow for other types of combination functions. For instance, an "N-of" combination function is the combination function for a binary effect variable which is equal to 1 if and only if $N$ or more binary mechanism variables are equal to 1. Clearly this can be generalized to handle continuous variables. By using such an additive threshold combination function one can capture threshholding effects in a causal interaction model. Another combination function is the X-or or parity combination function in which the binary effect variable is equal to 1 if and only if an even number of the binary mechanism variables are equal to 1. Causal interaction models with this combination function where the causes are jointly independent lead to a parameterized version of the pseudo-independence model of Xiang et al. (1996).

## 4 Learning the Structure and Parameters

In this section, we investigate how to learn the parameters and the structure for exponential causal interaction models. In Section 4.1, we show how to use the EM algorithm (Dempster et al., 1977) to compute the ML and MAP estimate of the parameters. In Section 4.2, we investigate asymptotic approximations of the marginal likelihood, in particular, the Cheeseman-Stutz approximation (1995).

### 4.1 Learning Parameters

We can write the causal interaction model as a DAG model. In particular, this means that we assume that the true (or physical) joint probability distribution for the set of variables $\mathbf{X} = \{X_1, \ldots, X_n\}$ in the DAG model can be encoded in some DAG model $S$. In this section, $X_1, \ldots, X_n$ are all of the variables in the model an not just the mechanism variables of a causal interaction model. We write

$$p(\mathbf{x}|\boldsymbol{\theta}_s, S) = \prod_{i=1}^{n} p(x_i|\mathbf{pa}_i, \boldsymbol{\theta}_i, S) \qquad (1)$$

where $\boldsymbol{\theta}_i$ is the vector of parameters for the distribution $p(x_i|\mathbf{pa}_i, \boldsymbol{\theta}_i, S)$, $\boldsymbol{\theta}_s$ is the vector of parameters $(\boldsymbol{\theta}_1, \ldots, \boldsymbol{\theta}_n)$. In addition, we assume that we have a random sample $D = \{\mathbf{x}_1, \ldots, \mathbf{x}_N\}$ from the true joint

probability distribution of **X**. We refer to an element $\mathbf{x}_l$ of $D$ as a *case*. Finally, we have a prior probability density function $p(\boldsymbol{\theta}_s|S)$ over the parameters of the DAG model. The problem of learning probabilities in a Bayesian network can now be stated simply: Given a random sample $D$, compute the posterior distribution $p(\boldsymbol{\theta}_s|D, S)$.

We refer to the conditional distribution $p(x_i|\mathbf{pa}_i, \boldsymbol{\theta}_i, S)$ as a *local (conditional) distribution function*. In this section, we illustrate the use of the EM algorithm in the case where each local distribution function is collection of multinomial distributions, one distribution for each configuration of $\mathbf{Pa}_i$. Namely, we assume

$$p(x_i^k|\mathbf{pa}_i^j, \boldsymbol{\theta}_i, S) = \theta_{ijk} > 0 \quad (2)$$

where $\mathbf{pa}_i^1, \ldots, \mathbf{pa}_i^{q_i}$ ($q_i = \prod_{X_i \in \mathbf{Pa}_i} r_i$) denote the configurations of $\mathbf{Pa}_i$, and $\boldsymbol{\theta}_i = ((\theta_{ijk})_{k=2}^{r_i})_{j=1}^{q_i}$ are the parameters. (The parameter $\theta_{ij1}$ is given by $1 - \sum_{k=2}^{r_i} \theta_{ijk}$.) For convenience, we define the vector of parameters

$$\boldsymbol{\theta}_{ij} = (\theta_{ij2}, \ldots, \theta_{ijr_i})$$

for all $i$ and $j$. We assume that each vector $\boldsymbol{\theta}_{ij}$ has the prior distribution $\mathrm{Dir}(\boldsymbol{\theta}_{ij}|\alpha_{ij1}, \ldots, \alpha_{ijr_i})$. $N_{ijk}$ is the number of cases in $D$ in which $X_i = x_i^k$ and $\mathbf{Pa}_i = \mathbf{pa}_i^j$.

Define $\tilde{\boldsymbol{\theta}}_s$ to be the configuration of $\boldsymbol{\theta}_s$ that maximizes $g(\boldsymbol{\theta}_s)$.

$$g(\boldsymbol{\theta}_s) \equiv \log(p(D|\boldsymbol{\theta}_s, S) \cdot p(\boldsymbol{\theta}_s|S)) \quad (3)$$

This configuration also maximizes $p(\boldsymbol{\theta}_s|D, S)$, and is known as the *maximum a posteriori* (MAP) configuration of $\boldsymbol{\theta}_s$. Also define $\hat{\boldsymbol{\theta}}_s$ to be the configuration of $\boldsymbol{\theta}_s$ that maximizes $p(D|\boldsymbol{\theta}_s, S)$. This configuration is known as the *maximum likelihood* (ML) configuration of $\boldsymbol{\theta}_s$.

In the case of causal interaction models, we need to compute the posterior given incomplete data. Unlike the complete-data case, we need to use approximation techniques. For more details see, for instance, Heckerman (1995). These techniques include Monte Carlo approaches such as Gibbs sampling and importance sampling (Neal, 1993; Madigan and Raftery, 1994), asymptotic approximations (Kass et al., 1988), and sequential updating methods (Spiegelhalter and Lauritzen, 1990; Cowell et al., 1995).

The asymptotic approximations are based on the observation that, as the number of cases increases, the posterior on the parameters will be distributed according to a multivariate-Gaussian distribution. As we continue to get more cases the Gaussian peak will become sharper, tending to a delta function at the MAP configuration $\tilde{\boldsymbol{\theta}}_s$. In this limit, we can use the MAP configuration to approximate the distribution.

A further approximation is based on the observation that, as the sample size increases, the effect of the prior $p(\boldsymbol{\theta}_s|S)$ diminishes. Thus, we can approximate $\tilde{\boldsymbol{\theta}}_s$ by the maximum *maximum likelihood* (ML) configuration of $\boldsymbol{\theta}_s$.

One class of techniques for finding a ML or MAP is gradient-based optimization. For example, we can use gradient ascent, where we follow the derivatives of $g(\boldsymbol{\theta}_s)$ or the likelihood $p(D|\boldsymbol{\theta}_s, S)$ to a local maximum. Russell et al. (1995) and Thiesson (1995) show how to compute the derivatives of the likelihood for a Bayesian network with unrestricted multinomial distributions. Buntine (1994) discusses the more general case where the likelihood function comes from the exponential family. Of course, these gradient-based methods find only local maxima.

Another technique for finding a local ML or MAP is the expectation–maximization (EM) algorithm (Dempster et al., 1977). To find a local MAP or ML, we begin by assigning a configuration to $\boldsymbol{\theta}_s$ somehow (e.g., at random). Next, we compute the *expected sufficient statistics* for a complete data set, where expectation is taken with respect to the joint distribution for **X** conditioned on the assigned configuration of $\boldsymbol{\theta}_s$ and the known data $D$. In our discrete example, we compute

$$\mathrm{E}_{p(\mathbf{x}|D, \boldsymbol{\theta}_s, S)}(N_{ijk}) = \sum_{l=1}^{N} p(x_i^k, \mathbf{pa}_i^j|\mathbf{y}_l, \boldsymbol{\theta}_s, S) \quad (4)$$

where $\mathbf{y}_l$ is the possibly incomplete $l$th case in $D$. When $X_i$ and all the variables in $\mathbf{Pa}_i$ are observed in case $\mathbf{x}_l$, the term for this case requires a trivial computation: it is either zero or one. Otherwise, we can use any Bayesian network inference algorithm to evaluate the term. This computation is called the *expectation step* of the EM algorithm.

Next, we use the expected sufficient statistics as if they were actual sufficient statistics from a complete random sample $D_c$. If we are doing an ML calculation, then we determine the configuration of $\boldsymbol{\theta}_s$ that maximize $p(D_c|\boldsymbol{\theta}_s, S)$. In our discrete example, we have

$$\theta_{ijk} = \frac{\mathrm{E}_{p(\mathbf{x}|D, \boldsymbol{\theta}_s, S)}(N_{ijk})}{\sum_{k=1}^{r_i} \mathrm{E}_{p(\mathbf{x}|D, \boldsymbol{\theta}_s, S)}(N_{ijk})}$$

If we are doing a MAP calculation, then we determine the configuration of $\boldsymbol{\theta}_s$ that maximizes $p(\boldsymbol{\theta}_s|D_c, S)$. In our discrete example, we have[2]

$$\theta_{ijk} = \frac{\alpha_{ijk} + \mathrm{E}_{p(\mathbf{x}|D, \boldsymbol{\theta}_s, S)}(N_{ijk})}{\sum_{k=1}^{r_i} (\alpha_{ijk} + \mathrm{E}_{p(\mathbf{x}|D, \boldsymbol{\theta}_s, S)}(N_{ijk}))}$$

---

[2]The MAP configuration $\tilde{\boldsymbol{\theta}}_s$ depends on the coordinate system in which the parameter variables are expressed.

This assignment is called the *maximization step* of the EM algorithm. Dempster et al. (1977) showed that, under certain regularity conditions, iteration of the expectation and maximization steps will converge to a local maximum. The EM algorithm is typically applied when sufficient statistics exist (i.e., when local distribution functions are in the exponential family), although generalizations of the EM have been used for more complicated local distributions (see, e.g., Saul et al., 1996).

## 4.2 Learning Structure

A key step in the Bayesian approach to learning graphical models is the computation of the marginal likelihood of a data set given a model $p(D|S)$. Given a *complete* data set—that is a data set in which each sample contains observations for every variable in the model, the marginal likelihood can be computed exactly and efficiently under certain assumptions (Cooper and Herskovits, 1992). In contrast, when observations are missing, including situations where some variables are *hidden* or never observed, the exact determination of the marginal likelihood is typically intractable. Consequently, we will use approximation techniques for computing the marginal likelihood of exponential causal interaction models.

In this section, we focus attentions on an asymptotic approximation called the Cheeseman-Stutz approximation, which use in the simulation study described in Section 5. It was chosen for the simulation study because of its computational and performance features. See Chickering and Heckerman (1996) for a discussion of other approximations and experimental results.

When computing most asymptotic approximations, we must determine the dimension of each of the model. The dimension of a model can be interpreted in two equivalent ways. First, it is the number of free parameters needed to represent the parameter space near the maximum likelihood value. Second, it is the rank of the Jacobian matrix of the transformation between the parameters of the network and the parameters of the observable (non-hidden) variables. In either case, the dimension depends on the value of $\hat{\boldsymbol{\theta}}_s$ space. In our simulation study we use a mathematical software package to calculate the rank of the Jacobian matrix of the transformation between the parameters of the network and the parameters of the observable variables. For more details and motivation see Geiger et al. (1996).

Now we turn our attention the the Cheeseman-Stutz approximation (1995). Recall that in the EM algorithm we treat expected sufficient statistics as if they are actual sufficient statistics. This use suggests an approximation for the marginal likelihood:

$$\log p(D|S) \approx \log p(D'|S) \qquad (5)$$

where $D'$ is an imaginary data set that is consistent with the expected sufficient statistics computed using an E step at a local ML value for $\boldsymbol{\theta}_s$.

Equation 5 has two desirable properties. One, because it computes a marginal likelihood, it punishes model complexity. Two, because $D'$ is a complete (albeit imaginary) data set, the computation of the criterion is efficient.

One problem with this scoring criterion is that it may not be asymptotically correct. Consider the asymptotically correct, Bayesian Information Criterion (BIC) (Schwarz, 1978; Haughton, 1988)

$$\log p(D'|S) = \log p(D'|\hat{\boldsymbol{\theta}}_s, S) - \frac{d'}{2} \log N + O(1)$$

where $d'$ is the dimension of the model $S$ given data $D'$ in the region around $\hat{\boldsymbol{\theta}}_s$—that is, the number of parameters of $S$. As $N$ increases, the difference between $p(D|\hat{\boldsymbol{\theta}}_s, S)$ and $p(D'|\hat{\boldsymbol{\theta}}_s, S)$ may increase. Also, as we have discussed, it may be that $d' > d$. In either case, Equation 5 will not be asymptotically correct. A simple modification to Equation 5 addresses these problems:

$$\begin{aligned}\log p(D|S) &\approx \log p(D'|S) \qquad (6)\\ &- \log p(D'|\hat{\boldsymbol{\theta}}_s, S) + \frac{d'}{2} \log N \\ &+ \log p(D|\hat{\boldsymbol{\theta}}_s, S) - \frac{d}{2} \log N\end{aligned}$$

Equation 5 (without the correction to dimension) was first proposed by Cheeseman and Stutz (1995) as a scoring criterion for AutoClass, an algorithm for data clustering. We shall refer to Equation 6 as the *Cheeseman-Stutz* scoring criterion. We note that the scoring crireria given in Equation 5 and Equation 6 can be applied if one can compute the marginal likelihood of complete data given the model and obtain a MAP estimate. Buntine (1994) shows how to compute the marginal likelihood for complete data given a DAG model in which the local likelihoods are from the exponential family and we will use the EM algorithm to obtain a MAP estimate.

## 5 Simulation Study

In this section we describe a small simulation study which highlights some of the important features of the approach that we described in Section 4. The structure of the five models that we used in the simulation study are given in Figure 4. All of the variables are binary and the conditional distributions of the $X_i$'s given the $C_i$'s are complete probability tables. Model F5 could also be represented without a deterministic combination function as a complete probability table of $E$ given the $C_i$'s.

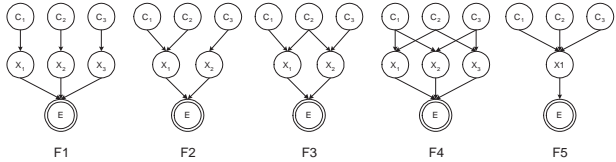

Figure 4: Noisy-Max-Interaction models used in simulation study.

| Model | F1 | F2 | F3 | F4 | F5 |
|---|---|---|---|---|---|
| Dimension | 7 | 7 | 9 | 10 | 11 |
| Unadjusted dim. | 9 | 9 | 11 | 15 | 11 |

Table 1: The dimension of the NMI models.

For each model we chose parameter values for the parameters and then used the parameterized model as the generating model to generate a dataset of 6400 cases. Parameter values were chosen by hand, however, similar results would be expected for parameters chosen at random. We approximated the model posteriors using the adjusted Cheeseman-Stutz score for different sized initial segments of the 6400 cases. The dimension of the models is calculated using Mathematica and the techniques described by Geiger et al. (1996). Although not done for our study, it is easy to automatically generate the equations for Mathematica to calculate the dimension and thus automate the calculation of dimension. The results of the simulation study are summarized in Figure 5. Model posteriors are presented only for initial segments of size 100, 200, 400, 800, and 1600 cases. Not surprisingly, mass continues to accumulate on the generating model as the sample size increases. The one exception is when model F1 is the generating model. The reason for the behavior of the posterior when F1 is the generating model is that the set of distributions that can be parameterized by F1 is a strict subset of the distributions that can be parameterized by F2 and, surprisingly, the dimension of the two models is identical. This unusual relationship between F1 and F2 only occurs only when the $C_i$'s and $E$ are binary.

Finally, we would like to draw attention to the importance of using the correct dimension when calculating the Bayesian approximation to the posterior. The *unadjusted dimension* of a DAG model is the number of parameters in the model, including the parameters for the hidden variables. Table 1 describes the dimension of each of the models used in the simulation study. Consider models F4 and F5. Clearly every distribution over the $C_i$'s and $E$ that can be represented in F4 can be represented in F5. If our asymptotic approximation used the unadjusted dimension then, at least asymptotically, it would be impossible to choose model F4 over model F5 when F4 is the generating model. Using the correct *penalty* for the dimension is also important for other approaches such as MDL.

## 6 Related and Future Work

There has be little work done on parameter learning for causal interaction models. The notable exception is the work of Neal (1992). Neal showed that one could learn the parameters of a noisy-or network using a local learning rule. However, his particular gradient-ascent procedure must be constrained to avoid entering an invalid region of the parameter space. Since we are using EM we are guaranteed to stay within the valid region of the parameter space and guaranteed to find a local maximum.

We plan on investigating the representational power of causal interaction models as compared to other local structures, e.g., decision graphs and compare the ease of assessment for various models In addition, we will consider automating the learning of causal interaction models (i.e., defining a search space, and search operators), and compare the result of such an algorithm to other approaches for learning local structure. Also of interest, is how to best combine a search for local structure with a search for global structure.

## Acknowledgments

We thank Bo Thiesson, Jack Breese, and Max Chickering for helpful discussion on this material.

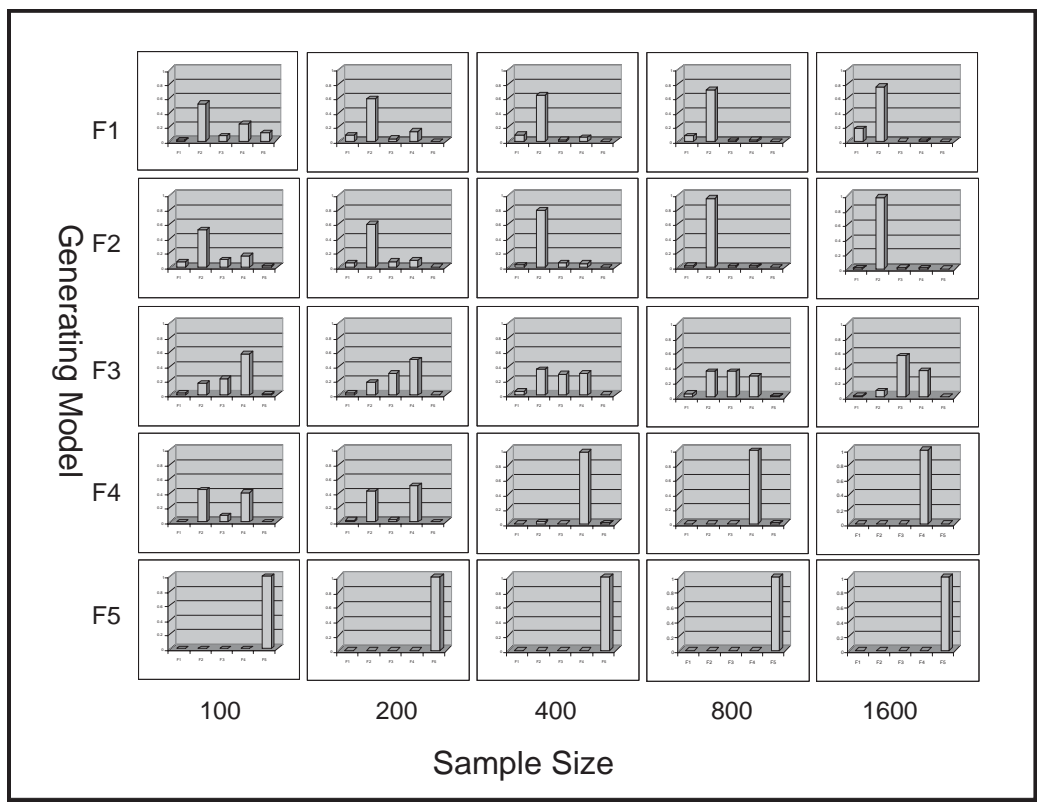

Figure 5: Simulation study results.